\documentclass[journal]{IEEEtran}
\usepackage{cite}
\usepackage{amsmath,amssymb,amsfonts}
\usepackage{algorithmic}
\usepackage{graphicx}
\usepackage{textcomp}
\usepackage[ruled,linesnumbered,vlined]{algorithm2e}
\usepackage{threeparttable}

\begin{document}

\title{Generalizable Industrial Visual Anomaly Detection with Self-Induction Vision Transformer}

\author{Haiming Yao,~\IEEEmembership{Student Member,~IEEE,} WenYong Yu,~\IEEEmembership{Member,~IEEE}


\thanks{Haiming Yao is with the State Key Laboratory of Precision Measurement Technology and Instruments, Department of Precision Instrument, Tsinghua University, Beijing 100084, China. (e-mail: yhm22@mails.tsinghua.edu.cn).}

\thanks{Wenyong Yu is with with the School of Mechanical Science and Engineering, Huazhong University of Science and Technology, Wuhan 430074, China (e-mail:ywy@hust.edu.cn).}
}
\markboth{Journal of \LaTeX\ Class Files,~Vol.~14, No.~8, August~2021}%
{Shell \MakeLowercase{\textit{et al.}}: A Sample Article Using IEEEtran.cls for IEEE Journals}

\maketitle
{}

\begin{abstract}
Industrial vision anomaly detection plays a critical role in the advanced intelligent manufacturing process, while some limitations still need to be addressed under such a context. First, existing reconstruction-based methods struggle with the identity mapping of trivial shortcuts where the reconstruction error gap is legible between the normal and abnormal samples, leading to inferior detection capabilities. Then, the previous studies mainly concentrated on the convolutional neural network (CNN) models that capture the local semantics of objects and neglect the global context, also resulting in inferior performance. Moreover, existing studies follow the individual learning fashion where the detection models are only capable of one category of the product while the generalizable detection for multiple categories has not been explored. To tackle the above limitations, we proposed a self-induction vision Transformer(SIVT) for unsupervised generalizable multi-category industrial visual anomaly detection and localization. The proposed SIVT first extracts discriminatory features from pre-trained CNN as property descriptors. Then, the self-induction vision Transformer is proposed to reconstruct the extracted features in a self-supervisory fashion, where the auxiliary induction tokens are additionally introduced to induct the semantics of the original signal. Finally, the abnormal properties can be detected using the semantic feature residual difference. We experimented with the SIVT on existing Mvtec AD benchmarks, the results reveal that the proposed method can advance state-of-the-art detection performance with an improvement of 2.8-6.3 in AUROC, and 3.3-7.6 in AP.

\end{abstract}


\section{Introduction}
\label{sec:introduction}
\IEEEPARstart{V}{isual} anomaly detection(VAD) has attracted increasing attention in the intelligent industrial field recently. As an automated, non-destructive, and efficient inspection technique, VAD has been deployed in many practical industrial scenarios including textured surface defect inspection\cite{r1}\cite{r2}, product quality inspections \cite{r3}, and smart power system\cite{r4}, etc, which can significantly improve the efficiency and accuracy of the inspection compared to the traditional manual inspection.

\begin{figure}[t]\centering
\includegraphics[width=8.8cm]{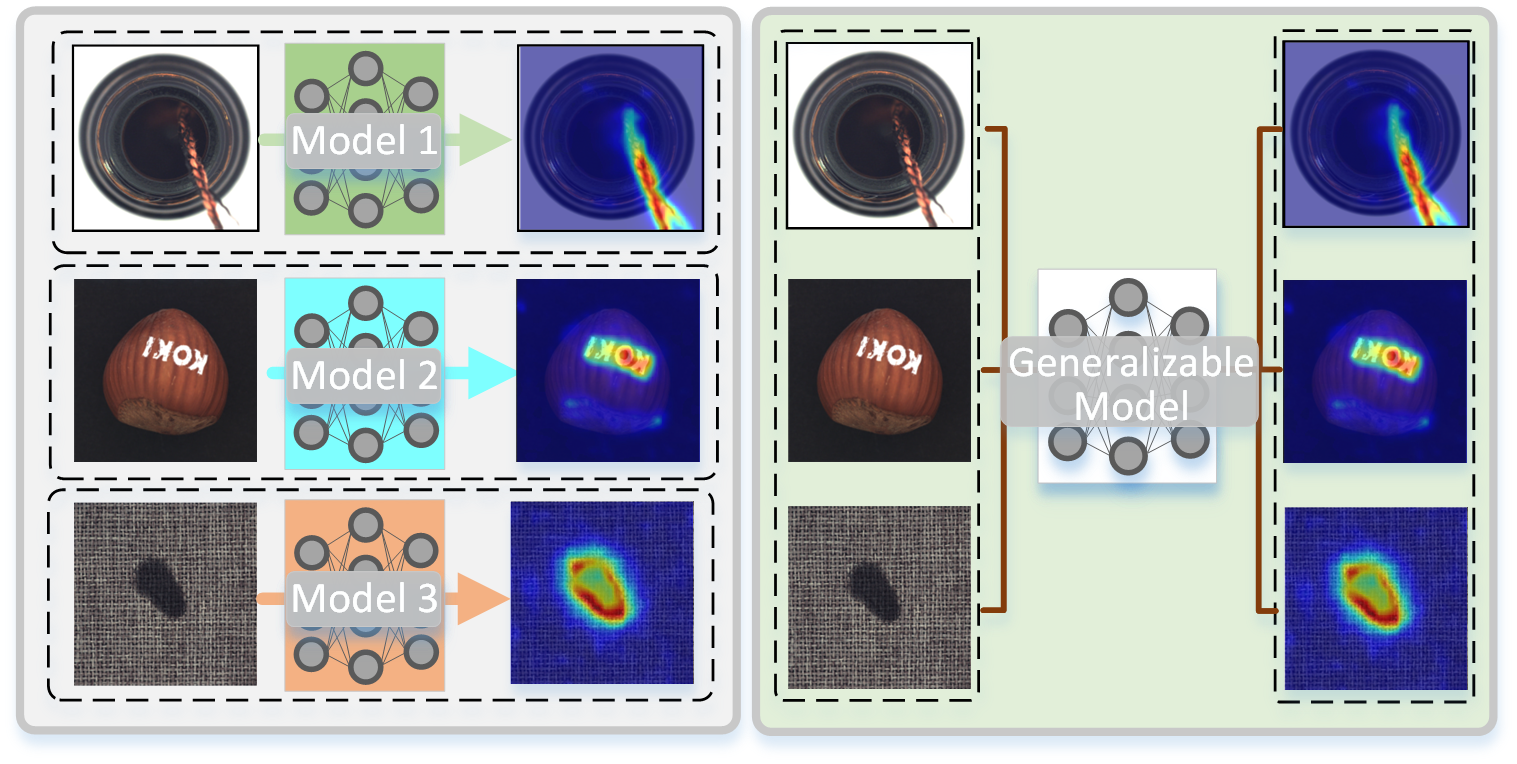}
\caption{Schema of existing individual detection paradigms versus challenging generalizable detection paradigms.}
\label{FIG0}
\end{figure}
The current methods for industrial VAD can be divided into two main categories: supervised learning-based and unsupervised learning-based approaches. Driven by a large number of defective samples and corresponding labels, the supervised learning-based approaches\cite{r4}\cite{r5} can yield accurate inspection results, however, it is impracticable and labor-intensive to obtain sufficient defect samples and labels in actual industrial settings. Thus, the unsupervised methods\cite{r1}\cite{r2}\cite{r3} that require only accessible defect-free samples have shown greater adaptability and have attracted more attention.

In the unsupervised VAD community, the reconstruction-based approaches are one of the most followed long-standing baseline models. The network model, which is usually named autoencoders(AEs), only adopts the normal samples in the training phase and leverages the reconstruction of input as the optimization objectives to model the distribution of normal patterns. Then, the reconstruction error gap between the normal and abnormal samples can be employed as the criterion for anomaly detection in the testing phase. This paradigm has been in-depth investigated. For example, in\cite{r6}, the MSCDAE is proposed to apply a three-layer pyramid denoising AE for textural surface defect detection. The AFEAN proposed in\cite{r2} employed anomaly feature editing to restore the latent abnormal feature. The FMR-Net\cite{r1} leverages the feature memory and rearrangement is also proposed for abnormal defects segmentation. However, most of the reconstruction-based approaches can still not obtain a satisfactory performance for complex cases due to the following limitations: (1). the over-generalizability of AEs results in the reconstruction behaviors of abnormal defects that are similar to that of the normal data. Therefore, a legible gap can not be obtained between the normal and abnormal samples. (2) the low-level per-pixel difference is semantically inferior, leading to an unreliable anomaly segmentation. (3) As illustrated in Fig. 1, the normality is modeled within the individual category in available methods while the challenging inter-category anomaly detection has not been developed yet.

Benefiting from bio-mimetic visual perception, CNNs have achieved remarkable performance in various downstream vision tasks. Among the current VAD methods, CNNs are also the mainstream network structure, which almost dominated all existing VAD methods. However, due to the limited receptive field, CNNs struggle to understand the global semantics, which causes the discriminability between the defective and normal patterns\cite{r7}. Recently, in the context of visual understanding, the vision Transformer is proposed in \cite{r8} to model the images from a sequence perspective using the self-attention mechanism, which can capture the global context and can be leveraged to overcome the limitations of CNNs in VAD.

Driven by the above analysis, leveraging the advantages of CNN and vision Transformer, we developed a hybrid reconstruction framework SIVT for generalizable cross-category industrial VAD. The TV first exploits the local semantic extraction capability of CNN to convert low-level image pixels into semantic deep feature maps. Then the self-induction vision Transformer is proposed by additionally introducing the auxiliary inductive tokens as the global semantic representation of the partial token subsets of the original feature in the latent space, thus casting the feature reconstruction as a self-introduction paradigm of the deep feature. The main contributions of this study are twofold: 
\begin{enumerate}
	\item A new hybrid VAD approach named self-induction vision Transformer(SIVT) framework is proposed for generalizable multi-category anomaly detection. The
SIVT consists of a feature extraction sub-network based on the pre-trained CNN and a feature reconstruction sub-network based on the vision Transformer. The feature extraction network is responsible for transforming low-level pixels into high-semantic feature maps, while the feature reconstruction sub-network is employed to reconstruct the anomalous features into normal patterns using the global contextual self-introduction paradigm. 

	\item We evaluate our SIVT method on the representative industrial benchmark MVTec AD dataset. The results demonstrate that SIVT achieves state-of-the-art performance with a comprehensive area under the receiver operating characteristic curve (AUROC) of 96.4 for anomaly detection and 96.9 for anomaly segmentation for all 15 categories under the generalizable detection paradigm, surpassing the existing method 2.8-5.3 in AUROC. 
 
\end{enumerate}

The remainder of this article is structured as follows: The related work associated with VAD, especially the reconstruction-based methods and vision Transformer is stated in Section II. The proposed SIVT methodology is detailed in Section III. In Section IV, the experimental results and analysis are illustrated. Finally, Section V concludes this article.

\section{Related work}
\label{sec: Related work}

There is a considerable amount of work focusing on the VAD, we specialize in the topic of reconstruction-based and vision Transformer-based approaches.

To address the limitations of vanilla AE, many extensions have been suggested to promote its capabilities. The MemAE proposed in \cite{r9} designs a memory bank to explicitly restore the normal latent representations and mitigate the over-generalization of vanilla AE. To consider structural properties, the SSIM loss is introduced in \cite{r13} which exhibits more sensitivity than the per-pixel difference. The Generative Adversarial Networks(GANs) are integrated with AE in \cite{r10}\cite{r11} to control the distribution of reconstructed images conforms to that of normal training data. The MLIR model in \cite{r12} is proposed to obtain flexible multilevel reconstruction results. The RIAD model was proposed in \cite{r16}, which converts the reconstruction into multiple partial images inpainting. The above methods are all conducted based on the pixel reconstruction difference. The feature reconstruction difference is first proposed in DFR\cite{r14}, DFR employed a pre-train VGG19 network to extract hierarchical features as the substitution for image pixels, which demonstrates a much more robust anomaly detection performance. A dual-Siamese network proposed in \cite{r15} extra employs the synthetic defects to obtain a more stable reconstruction effect. As the generalized reconstruction schemes, the concept of feature space mapping and correspondence are proposed in \cite{r17} and \cite{r18}, in which the deep features from two spaces are bridged together for normal patterns while being separated for abnormal patterns. In general, though the above reconstruction-based methods can be effective for the simple defect categories, some challenging types of defects\cite{r17} are difficult to solve due to the inherent limitations of CNNs. Moreover, the above works are designed under the one-model-per-category learning fashion and do not consider cross-category generality.

As a new genre of architecture in the field of vision, the vision Transformer has been explored and proven to be efficient for a wide range of visual tasks\cite{r22}. However, in the CNN-dominated VAD community, the vision Transformer has not been sufficiently researched yet. The Intra in \cite{r23}  poses the image reconstruction as a patch-inpainting problem. In \cite{r19}, the U-TRansformer that constructed a reconstruction network with a pyramidal hierarchy of skip connections is introduced for VAD. The MSTUnet\cite{r20} which developed an inpainting Swim Transformer and discriminate network denotes precise anomaly localization effect. ST-MAE is proposed in \cite{r21} that is based on the MAE\cite{r22} utilizing the Siamese encoder structure and paradigm of deep feature transition for uniform VAD in multiple application scenarios. However, compared with the relatively established CNN-based VAD methods, there is still substantial potential for the development of vision Transformer-based VAD models. 

In this paper, we proposed a robust reconstruction method namely self-induction vision Transformer(SIVT) in this study to circumvent the limitations of the existing reconstruction-based approaches and to pioneer generalizable multi-category detection paradigms.

\section{Proposed SIVT Methodology}

\begin{figure*}[t]
\centerline{\includegraphics[width=\textwidth]{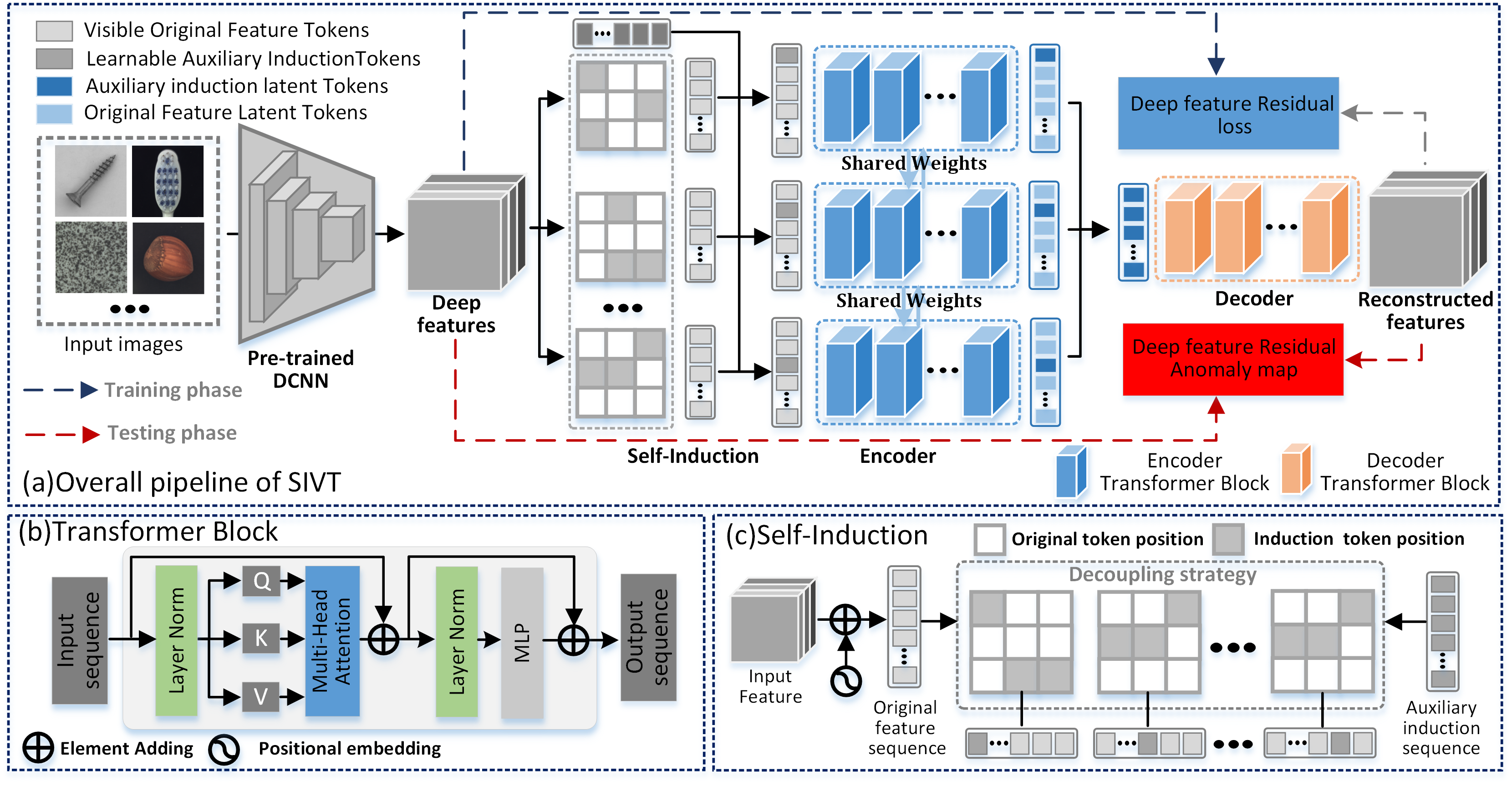}}
\caption[width=\textwidth]{
(a) The overall pipeline of the proposed SIVT. The SIVT firstly leverages pre-trained DCNN to extract deep feature of input images, and then introduce the self-induction mechanism enabled vision Transformer to reconstruct the deep features. (b) The typical self-attentive computational process in Vision Transformer \cite{r8}. (c) The self-induction Schematic.
}
\label{fig2}
\end{figure*}

\subsection{Overall pipeline}
The overall framework of the proposed SIVT is introduced in Fig. 1. Our SIVT follows a hybrid design that incorporates a CNN and a vision Transformer. The input images are first propagated into a pre-trained CNN to convert image pixels into semantic representational feature maps. Then, the self-induction vision Transformer based on the encoder-decoder architecture is introduced for feature reconstruction. In SIVT, the feature map is first transformed into a feature patch token sequence and divided into multiple disjoint partial subsets. Then each subset is additionally introduced with the learnable auxiliary induction tokens and subsequently fed into a Transformer encoder, in which the semantic information of each subset is aggregated into the latent auxiliary token. Next, the auxiliary induction tokens in each latent sequence are re-assembled into a single entire sequence while the other tokens from the original sequence are all discarded. Afterward, the re-assembled sequence is processed by a Transformer decoder to reconstruct the original feature. Finally, the feature reconstruction residual difference can be used for robust anomaly detection and segmentation. In the following, our method is illustrated in detail.

\subsection{Feature extraction by pre-trained CNN}
Instead of directly using the image pixel as the reconstruction objective, we adopt a more robust and semantic property representation for the image, where the CNN is adopted as the discriminative feature extractor. Specifically, given the input image $\mathcal{I}\in \mathbb{R} ^{H\times W\times 3}$, it is fed into the fixed CNN model pre-trained on the ImageNet dataset. The hierarchical features from different convolutional blocks are selected and adjusted to the same spatial size, then concatenated in the channel dimension to form fused multi-scale deep feature maps $\mathcal{F}\in \mathbb{R} ^{\mathcal{H} \times \mathcal{W}\times \mathcal{C}}$, which is equipped with sufficient discriminability hierarchical local semantics.

Compared with the low-level pixel representation, the deep feature maps $\mathcal{F}$ contain rich receptive information
and has more local contextual semantics, which is much beneficial in anomaly pattern revealing.

\subsection{Feature reconstruction by SIVT}

With the extracted deep feature maps $\mathcal{F}$. The SIVT is introduced to perform the feature reconstruction. Unlike the standard ViT \cite{r8}, we first divide the $\mathcal{F}$ into regular non-overlapping feature patches with the size of $P$ to reduce the input sequence length for the Transformer with a more acceptable computation consumption. Each patch is therefore a square of $P\times P$ feature vectors, which is then mapped into $D$-dimentional tokens and added to the corresponding position embeddings to construct the original feature patch token sequence $\mathcal{X} \in \mathbb{R}^{L\times D}$, where $L=\frac{\mathbf{H}}{P} \times \frac{\mathbf{W}}{P}$.

In this paper, we proposed the self-induction paradigm for feature patch token sequence reconstruction. Concretely, we first introduce the learnable auxiliary induction sequence $\mathcal{X}^{*}$, which has the same size as $\mathcal{X}$. Then, we developed a decoupling strategy to randomly partition the $\mathcal{X}^{*}$ into $N$ disjoint subsets $(\mathcal{X}_{1}^{*},\mathcal{X}_{2}^{*},...,\mathcal{X}_{N}^{*})$, each containing $\frac{1}{N}$ fraction of total tokens. On this basis, we then further constructed a set of sequences $\mathbf{\Phi} =(\Phi_{1}, \Phi_{2},...,\Phi_{N})$, which subjects to the following condition:
\begin{equation}
\mathbf{\Phi} =\left \{\Phi_{1},\Phi_{2},...,\Phi_{N}\mid \Phi_{i}=\mathcal{X}_{i}^{*}\cup \complement_\mathcal{X}\mathcal{X}_{i}^{*}, i=1,...,N  \right \} 
\end{equation}
where $\complement_\mathcal{X}\mathcal{X}_{i}^{*}$ denotes the part of tokens in $\mathcal{X}$ that is complementary to the $\mathcal{X}_{i}^{*}$ in terms of position. Thus, each sequence $\Phi_{i}$ consists of $\frac{1}{N}$ fraction of learnable auxiliary induction tokens and $\frac{N-1}{N} $ proportion of original feature patch tokens.

Subsequently, the $\mathbf{\Phi}$ is fed into the Transformer encoder to obtain a set of latent representation sequences $\mathbf{Z}_{\mathbf{\Phi}}=\left \{ Z_{\Phi_{1}}, Z_{\Phi_{2}},..., Z_{\Phi_{N}}  \right \} $. In this process, the learnable auxiliary induction tokens $\mathcal{X}_{i}^{*}$ act as the aggregating representations which induct the semantics of the remaining original feature patch tokens. Next, after encoding, we discard the latent representation of all the original feature patch tokens and retain that of the auxiliary induction tokens for each sequence, which are re-assembled into a single entire latent induction sequence, denoted as $Z_{\mathcal{X}^{*}}$, which is finally processed by the Transformer decoder to reconstruct the original feature.

Notably, most of the existing reconstruction-based approaches follow the autoencoding framework, where the input is encoded into the latent representations and decoded to the original signal, leading to the drawbacks of identity mapping. In contrast to that, our self-induction method generates the latent representations by additional introducing the learnable auxiliary induction sequence, which inducts the semantics of the original signal. Since the induction tokens only aggregate the semantics of the original tokens that do not overlap with them in the encoding phase, each latent embedding of the induction token in the latent sequences is generated by conditioning only the information of its disjoint original feature token subset, and thus semantically plausible, avoiding the trivial anomaly reconstruction phenomenon observed in traditional reconstruction approaches. Moreover, the proposed self-induction can be viewed as an implicit self-supervisory task that encourages the Transformer encoder to learn a holistic understanding of object semantics, which is very beneficial to reconstruction-based VAD approaches. 

\begin{algorithm}[t]
\SetAlgoLined
\caption{Feature reconstruction by self-induction}
 \SetKwInOut{Input}{input}
\SetKwInOut{KwOut}{Parameters}
\KwIn{Input feature $\mathcal{F}$}
 \KwOut{Patch size $P$; Subset number $N$}
\KwResult{Reconstruction feature $\tilde{\mathcal{F}}$}

Embed the $\mathcal{F}$ into sequence: $\mathcal{X}=\mathbf{E}(\mathcal{F},P)$\;
Given the auxiliary induction sequence $\mathcal{X}^{*}$\; 
Randomly divide $\mathcal{X}^{*}$ into $N$ disjoint token subsets: $\Big \{\mathcal{X}^{*}_{i}, i= \left \{ 1,..,N \right \} \mid \bigcup_{i}^{N}\mathcal{X}^{*}_{i}= \mathcal{X}^{*} \Big \} =D(\mathcal{X}^{*},N)$\;
\For{$i=1,...,{N}$}
{
Replace the corresponding tokens in $\mathcal{X}$ with $\mathcal{X}_{i}^{*}$:
	$\Phi_{i}=\mathcal{X}_{i}^{*}\cup \complement_\mathcal{X}\mathcal{X}_{i}^{*}$\;

	}
Encode the $\mathbf{\Phi} =\left \{ \Phi_{i}, i=\left \{ 1,...,N \right \} \right \}$ with $\mathbf{Enc}$:
$\mathbf{Z}_{\mathbf{\Phi}}=\left \{ Z_{\Phi_{1}}, Z_{\Phi_{2}},..., Z_{\Phi_{N}}  \right \} = \mathbf{Enc}\left ( \mathbf{\Phi} \right ) $ \;

\For{$i=1,...,{N}$}
{
Retain the induction tokens embedding in $Z_{\Phi_{i}}$:
	 $Z_{\mathcal{X}^{*}}=Z_{\mathcal{X}^{*}}\cup Z_{\Phi_{i}}\left ( \mathcal{X}^{*}_{i} \right ) $ \;
	}
Decode the $Z_{\mathcal{X}^{*}}$ with $\mathbf{Dec}$:
$\tilde{\mathcal{F}} = \mathbf{Dec}\left ( Z_{\mathcal{X}^{*}}\right )$\;

return $\tilde{\mathcal{F}}$
\end{algorithm}






The feature reconstruction by self-induction procedure is written more compactly in Algorithms 1.

\subsection{Objective function for training}

The Euclidean distance loss and the cosine similarity loss are adopted for model training. Given the input feature $\mathcal{F}\in \mathbb{R} ^{\mathcal{H} \times \mathcal{W}\times \mathcal{C}}$, the SIVT reconstructs it as $\tilde{ \mathcal{F}}\in \mathbb{R} ^{\mathcal{H} \times \mathcal{W}\times \mathcal{C}}$, the reconstruction loss is as follows:
\begin{equation}
\ell =\frac{1}{\mathcal{H}\times\mathcal{W}}\left\{\left ( \left \| \mathcal{F}-\tilde{\mathcal{F}}  \right \|_{2}^{2} \right) +\lambda\times  \left(   1-\frac{ \mathcal{F}\cdot \tilde{\mathcal{F}}}{\big\|\mathcal{F} \big\| \times \big\|\tilde{\mathcal{F}}\big\| }  \right)\right\} 
\end{equation}
where the $\lambda$ is the parameter that balances the relative contributions of the two loss terms.
\subsection{Anomaly detection for inference}
After training, the SIVT can be deployed for online visual anomaly detection. The normal features will be well reconstructed, while the reconstruction of abnormal features will be restrained since the SIVT only learns to induce anomaly-free
representations. Thus the feature reconstruction difference is assigned as the anomaly score map $\mathcal{A}\in \mathbb{R}^{\mathcal{H}\times \mathcal{W}}$:
\begin{equation}
\mathcal{A}  = \left \| \mathcal{F}-\tilde{\mathcal{F}}  \right \|_{2}^{2} \otimes \big(  1-\frac{ \mathcal{F}\cdot \tilde{\mathcal{F}}}{\big\|\mathcal{F} \big\|\times   \big\|\tilde{\mathcal{F}}\big\| }  \big)
\end{equation}
where the $\otimes$ is the element multiplication. Finally, the spatial resolution of the anomaly map is changed to the input image size using the interpolation operation, and A Gaussian filter is employed to smooth the final anomaly map. The standard deviation of $\mathcal{A}$ is adopted as the image-level anomaly score.

\section{Experimental results}
\label{sec:Experimental}
In this section, we validate our SIVT approach on the Mvtec AD\cite{r24} benchmark. With 5 categories of textures and 10 categories of objects, the Mvtec AD is composed of 3629 normal images as the training set; 467 normal images, and 1258 abnormal images as the testing set. The abnormal defect types are abundant and representative of real industrial scenarios. Unlike existing methods, we conducted experiments in the generalizable running mode, where all 15 categories are trained simultaneously to obtain a generalizable model.
Following the customary practice in previous works, the area under the receiver operating characteristic (AUROC) and average precision(AP) are utilized as the quantitative evaluation indicators for both image and pixel-level anomaly detection.

\subsection{Implementation Setup}

Each image is reshaped into the resolution of 256×256 and is normalized by the standard deviation and mean value obtained on the ImageNet dataset. The pre-trained  EfficientNet-b4 CNN model is exploited as the feature extractor, and the multi-scale feature maps are extracted from the intermediate outputs of the first for convolutional modules, which are resized into the same spatial size of 32 and concatenated in the channel dimension to form a $\mathcal{F} \in \mathbb{R} ^{32\times 32\times 272}$ tensor. The patch size $P$ and the subset number $N$ are set to 2 and 4 by default. The structural configuration of SIVT follows the design of the MAE-base but we reduce the embedding dimension to 240 for efficient computation. The SIVT model is trained from scratch by an AdamW optimizer with a learning rate of 1e-4 and batch size of 8 for 400 training epochs. All the experiments are implemented on a desktop computer with Xeon(R) Gold 6226R CPU@2.90GHZ and NVIDIA A100 GPU with 80GB memory size.
\begin{table*}
\caption{{Image-level/pixel-level AUC ROC results of different methods in MVTec AD}}
\label{table}
\setlength{\tabcolsep}{3pt}
\begin{threeparttable}
\begin{tabular}{p{\textwidth}}
$\includegraphics[width=\textwidth]{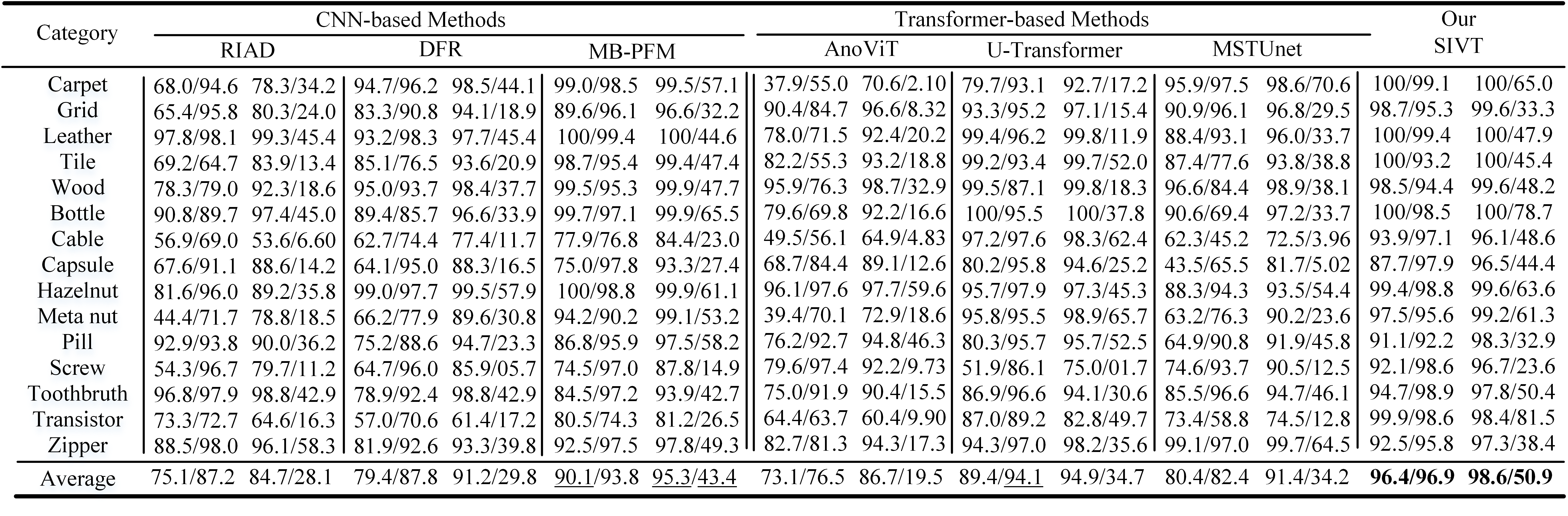}$
\end{tabular}
\begin{tablenotes}
       \footnotesize
       \item[1]The best AUC ROC performance is indicated by bold font, while the second best is indicated by an underline.
       \item[2]The results of each method are organized as follows: image/pixel AUROC; image/pixel AP.
\end{tablenotes}
\end{threeparttable}
\label{table4}
\end{table*}

\subsection{Comparative Experiments}

The proposed SIVT is compared with the existing cutting-edge reconstruction-based approaches, of which the RIAD \cite{r16}, DFR\cite{r14}, and MB-PFM\cite{r18} are schemes using CNN, while the AnoViT\cite{r25}, U-Transformer\cite{r19}, and MSTUet\cite{r20} are the vision Transformer enabled. For CNN models, the RIAD cast the image reconstruction as the inpainting problem to avoid the defects reconstruction; the image reconstruction is first promoted to feature reconstruction in DFR which can be viewed as a baseline for subsequent methods; the MB-FPM, on the other hand, is the most recent state-of-the-art model using the pre-trained feature sub-space mapping. In terms of vision transformer-based models, the Transformer encoder is introduced in AnoViT adopts to capture the global semantics in the image reconstruction; the U-Transformer is built with the pyramidal hierarchical structure with skip connections for feature reconstruction; while the MSTUet introduced the Swim Transformer as the reconstruction sub-network and CNN as the discriminative sub-network for end-to-end anomaly detection. To achieve a fair comparison, all the competitive approaches were trained under the challenging generalizable setting, in which all the 15 categories in MvtecAD are trained altogether without any categorical labels accessible during training and fine-tuning during testing.

\subsubsection{Quantitative Evaluation of the SIVT}
The quantitative experimental results are described in Table I. SIVT achieves the comprehensive 96.4/96.9 of image-/pixel-level AUROC, and 98.6/50.9 of image-/pixel-level AP in all 15 categories, significantly surpassing the competing methods. Compared with the sub-optimal methods MB-PFM and U-Transformer, SIVT improved the image/pixel-level AUROC by a margin of 6.3 and 2.8 and improved the image/pixel-level AP by 3.3 and 7.5. Among all categories, SIVT achieved the best or second-best accuracy in 13 categories of pixel-wise defect localization and 13 categories of image-wise defect detection. Though the performance gains are not much more than the MB-PFM and U-Transformer for homogenous texture categories, the SIVT is considerably superior to them for more sophisticated object categories. For example, for the most challenging category transistor, the SIVT can obtain remarkable performance improvements by margins of 12.9, 9.4, 15.6, and 31.8 for four indicators.

\begin{table*}
\caption{{The comparative results for the model generalizable robustness, efficiency, and consumption.}}
\label{table}
\setlength{\tabcolsep}{3pt}
\begin{threeparttable}
\begin{tabular}{p{\textwidth}}
$\includegraphics[width=\textwidth]{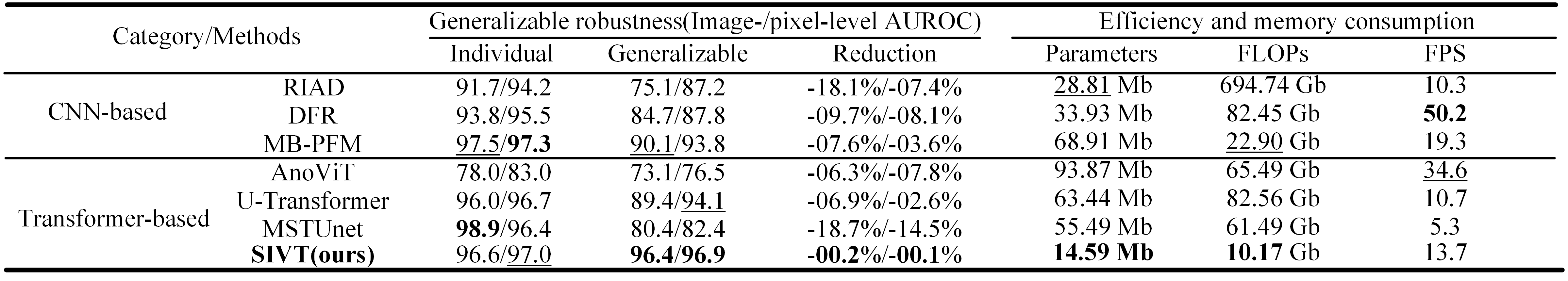}$
\end{tabular}

\end{threeparttable}
\label{table4}
\end{table*}
\begin{figure*}[t]
\centerline{\includegraphics[width=\textwidth]{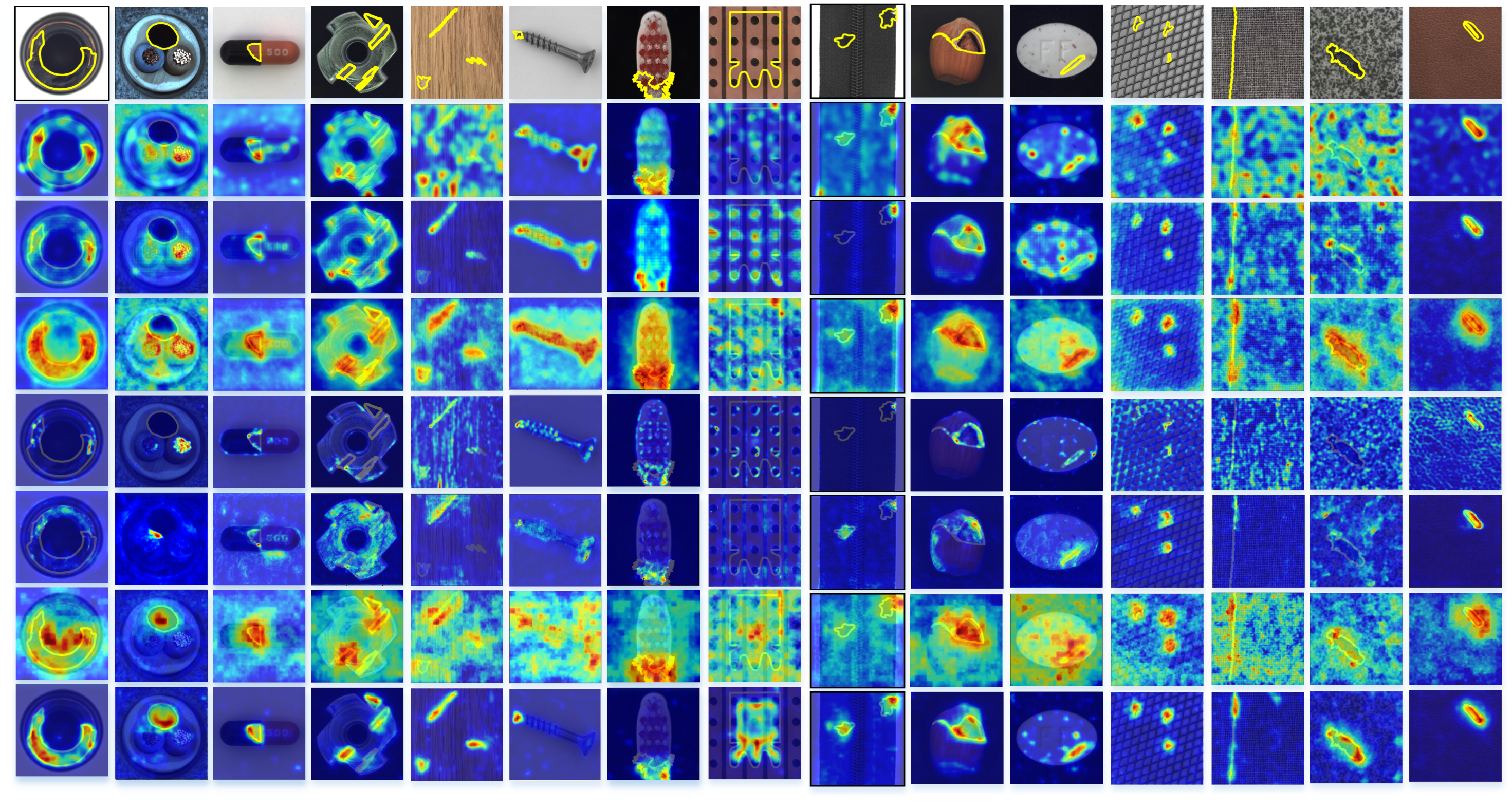}}
\caption[width=\textwidth]{
 Anomaly localization results of various samples in the Mvtec AD dataset with different methods. Top to bottom is the original defect samples where the ground truths are indicated by the yellow contours. The localization results were obtained by the RIAD, DFR, MB-PFM, AnoViT, U-Transformer, and the proposed SIVT model.
}
\label{fig2}
\end{figure*}

\subsubsection{Qualitative Evaluation of the SIVT}
As shown in Fig. 3, the SIVT is visually compared with the competing methods on the MVTec dataset.  In general, the SIVT can obtain the most accurate abnormality localization map. Especially for the categories with large variations, such as the cable, metal nut, screw, capsule, and transistor, the SIVT performs much better results. Note that the CNN-based models perform poorly in the above categories, this phenomenon occurs because the local receptive field of CNNs is restricted to tackle the defect types with the overall semantic drift, such as "misplaced" in the transistor, and 'missing cable' in the cable, which is improved considerably in the Transformer-based model. Among the competing methods, MSTUnet adopts the discriminative network for anomaly localization can obtain relatively precise results but will cause missed inspections where the defective regions are falsely identified as background; MB-PFM and U-Transformer can locate most defective areas but at the cost of noise injection in the background. Overall, our method can successfully accurately inspect various defects simultaneously whether they are local structural corruption or global semantic drift.

\subsubsection{Model generalizable robustness evaluation}
Generalizable running mode is a novel strategy, which enables the model to accomplish the detection task of multiple categories of anomalous defects using a generalizable framework. Compared to conventional solutions that are usually only applicable to the detection of a specific type of defect, it is challenging but practical in industrial scenarios.

As shown in Table II, the comparative experiments on the model's generalizable robustness were performed. All the compared models were carried out separately for the two paradigms where the 'Individual' denotes the conventional model corresponding to one category of products, the 'Generalizable' indicates the challenging one model applicable to all classes of products. Under the individual running mode, the proposed SIVT method does not outperform the MB-PFM and MSTUnet methods but still gets fairish results. Moreover, 
when switching to the generalizable running mode, the performances of all comparative model drop dramatically. The previously state-of-the-art(SOTA) CNN-based MB-PFM suffered a relative reduction of 7.6 $\%$ and 3.6 $\%$ in the image-/pixel-level AUROC. The advanced Transformer-based U-Transformer using the feature reconstruction resulted in a relative drop of 6.9$\%$ and 2.6$\%$ in image/pixel-level AUROC, while for the MSTUnet employ the pseudo-anomaly and discriminative sub-network, the drop is as large as 18.7$\%$ /14.5$\%$. In contrast, our SIVT has almost no performance degradation from individual modes to generalizable modes and surpasses the compared models dramatically by a dramatically large margin, demonstrating our superiority.

\subsubsection{Efficiency and consumption evaluation}

In addition to the performance, the model inference efficiency and the memory consumption are also the key indicators to evaluate the model. In this section, the model parameters, FLOPs(floating point operations per
second), and FPS(frame per second) are adopted to describe the model efficiency and consumption. 

The comparative results of different methods are displayed in TABLE II. It is noted that the FPS of the SIVT is 13.7, which is only behind the DFR, MB-PFM, and AnoviT, while the detection performance markedly surpasses them and can still meet the requirement of industrial deployment. Furthermore, in terms of memory consumption, The SIVT only takes 14.59 M of parameters and 10.17 G FLOPs, slighter than SOTA approaches MB-PFM and U-Transformer. Although other competitive methods adopt more complicated structure designs, they suffer more severe performance degradation when applying to the generalizable running mode, while our SIVT reveals stronger generalizable ability.

\subsection{Ablation Studies}
\subsubsection{Influence of the feature extraction}
Instead of image reconstruction, we performed feature extraction as a predecessor process to convert image reconstruction into feature reconstruction for more robust detection. The effectiveness of this method is experimentally studied in this section. As described in Fig. 4, different schemes are involved in the comparison including the image reconstruction and different CNN types\cite{r26} (Mobile Net, VGG19, ResNet34, and Efficientnet-b4). Overall, the feature reconstruction is much better than the image reconstruction and the best result was obtained with Efficientnet-b4. This result can be interpreted as the deeper features yield higher discriminability than the image pixel. Moreover, different networks have different semantic representation capabilities, with the best discriminative ability to be aware of the defective structural damages, the Efficientnet-b4 can work best with SIVT.

\begin{figure}[t]\centering
\includegraphics[width=8.8cm]{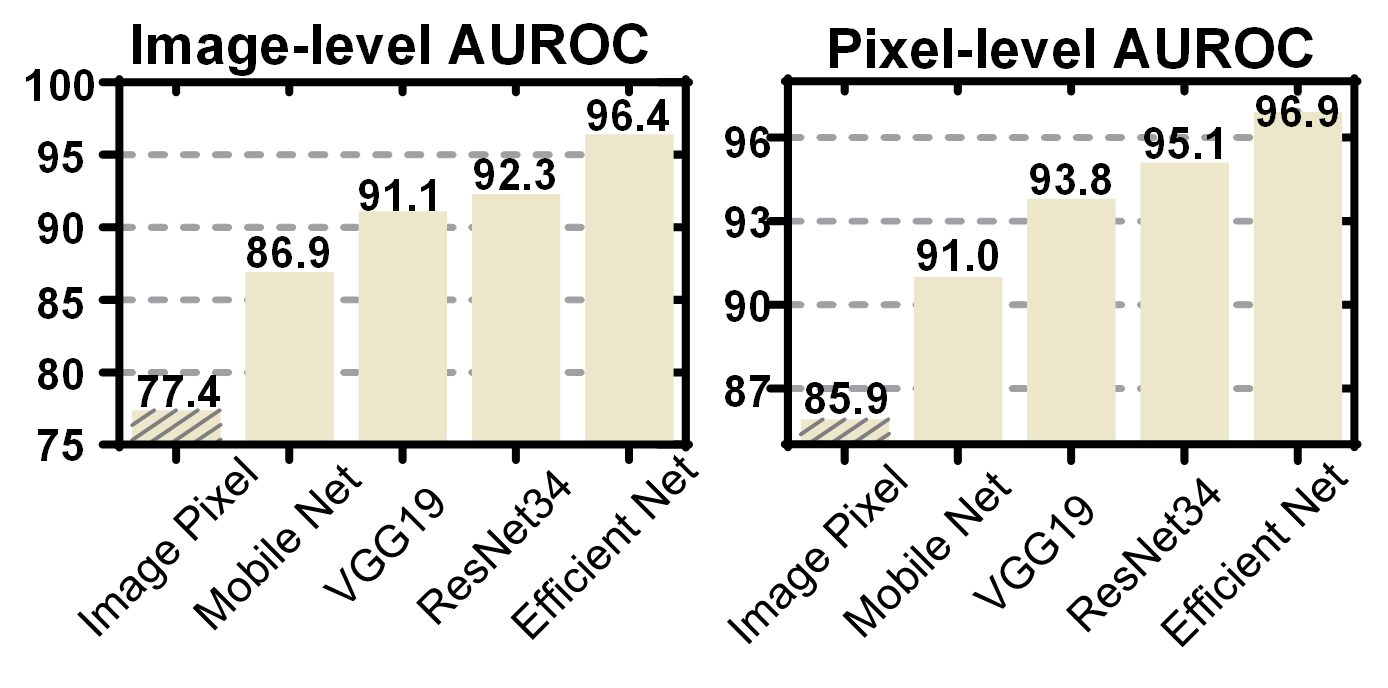}
\caption{The influence of the feature extraction on model performance.}
\label{FIG_1}
\end{figure}

\begin{figure}[t]\centering
\includegraphics[width=8.8cm]{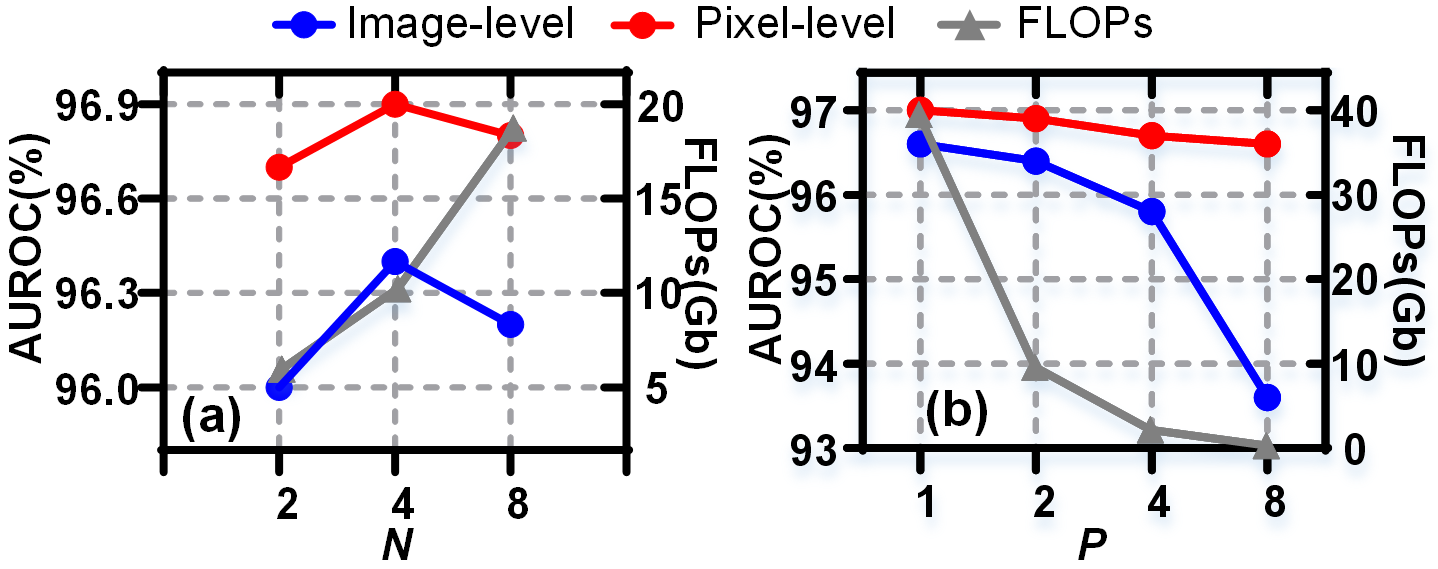}
\caption{(a) Influence of the number of subset number $N$. (b) Influence of the number of patch size $P$.}
\label{FIG_1}
\end{figure}

\begin{figure}[t]\centering
\includegraphics[width=8.8cm]{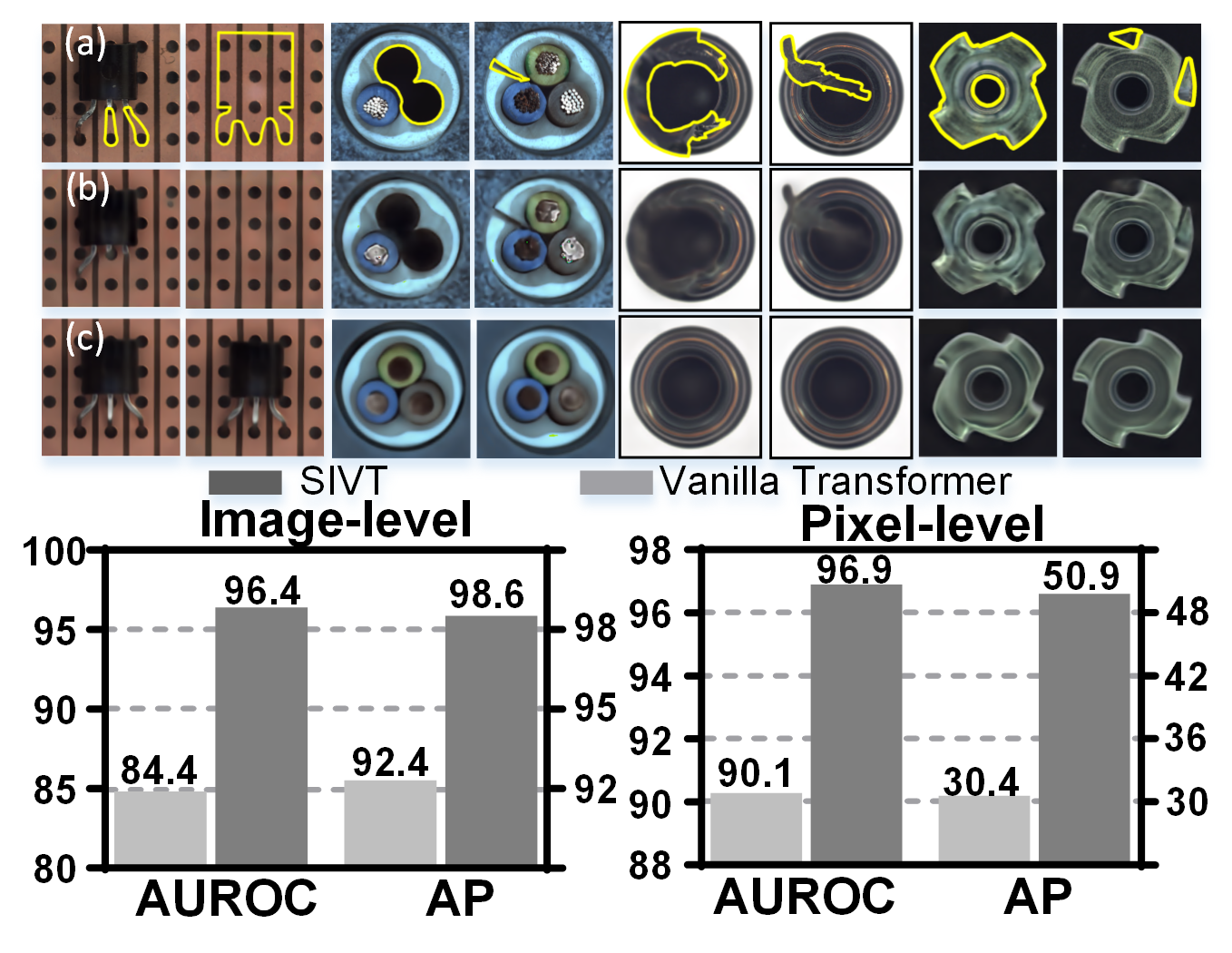}
\caption{Effect of the feature reconstruction. Upper part:(a). Original defective samples. (b). Reconstruction results by the vanilla Transformer. (c). Reconstruction results by the SIVT. Lower part: the quantitative results of the vanilla Transformer and SIVT.}
\label{FIG_1}
\end{figure}
\subsubsection{Impact of the feature reconstruction}
As stated before, the auxiliary induction mechanism is a crucial part that can convert the feature reconstruction into self-induction to prevent the trivial anomaly reconstruction phenomenon. A comparison between the vanilla Transformer and the SIVT is performed for feature reconstruction. Concretely, the vanilla Transformer adopts the same structure as SIVT for a fair comparison but directly reconstructs the input extracted feature following the encoding and decoding procedure.

First, the reconstruction results are visualized. For an intuitive verification, we extra employ a convolutional decoder to render the reconstructed features back to the image. The experimental results in the upper part of Fig. 6 show examples of rendered images of reconstructed features by the vanilla Transformer and SIVT. It can be viewed that the defective samples are reconstructed as normal patterns by SIVT while are well recovered by vanilla Transformer.

To analyze it quantitatively,  As indicated in the lower part of Fig. 6, the SIVT with self-induction substantially outperforms the vanilla Transformer based reconstruction model, the self-induction helps to improve the image-/pixel-level AUROC by margins of 12.0 and 6.8, and image-/pixel-level AP by margins of 6.2 and 20.5. 

The above experimental results strongly demonstrate that the self-induction mechanism is a significant module that can avoid trivial shortcuts and improve the performance of typical reconstruction-based VAD systems.


\subsubsection{Effect of the subset number}

As described in Section III.(C), the learnable auxiliary induction sequence $\mathcal{X}^{*}$ is decoupled into $N$ disjoint subsets to induce the semantics of complementary extracted feature patch token subsets. Here, we attempt to evaluate how the number of subsets will affects the model performance and efficiency.

Fig. 6 shows the image-/pixel-level AUROC and FLOPs with different $N$ values. It is clear to us that the FLOPs increases monotonically with $N$, while the detection performance shows a parabolic trend, rising first and then falling as $N$ varies from 2 to 8. This result occurs because a small $N$ value implies that there is a larger proportion of learnable auxiliary induction tokens in the input sequences $\mathbf{\Phi}$, retaining too little information of the original signal; However, a small number of learnable auxiliary induction tokens will be unable to induce the information of original signal with a too-large $N$ value, both of these cases result in poor reconstruction accuracy and thus performance degradation. Thus, $N$ is set to 4 accordingly.

\subsubsection{Sensitivity of the patch size}

Unlike the traditional hybrid CNN-Transformer architecture in \cite{r8} that directly applies the embedding projection on each vector in the feature map, we explored the feature patch embedding which can reduce the computational complexity and enable more effective industrial applications. Verification experiments are conducted in this section to investigate the influence of feature patch size $P$ on the model performance and efficiency.

The experimental results are displayed in Fig. 5(b), it can be seen that when $P$ vary in the range of $\left \{ 1,2,4,8 \right\}$, the image-/pixel-level AUROC first remains stable and then drops rapidly as $P$ increases further. Furthermore, the FLOPs values show a monotonic ascending trend with the increasing $P$. This is mainly because with the larger patch size, the total number of tokens will decrease. Typically, representing the extracted feature with more tokens contributes to higher reconstruction accuracy, but at the cost of a more intensive computational cost. Thus, to obtain a trade-off between performance and efficiency, the setting $P=2$ was chosen.

\section{Conclusion}
In this paper, we proposed a novel framework SIVT for the challenging generalizable industrial visual anomaly detection of multi-category products. The SIVT is a hybrid framework that adopts the CNN to extract local discriminatory features and leverage the global semantic capture capability of the vision Transformer to reconstruct the deep feature. Most importantly, the self-induction mechanism is proposed to augment the reconstruction-based VAD methods. We verified the efficiency of the proposed scheme through experiments, demonstrating its superior both in performance and efficiency. In the future, we will continually improve this method and tackle more practical and challenging VAD problems in the industrial field.

\bibliographystyle{ieeetr} 
\bibliography{reference}

\begin{thebibliography}{10}

\bibitem{r1}
H.~Yao, W.~Yu, and X.~Wang, ``A feature memory rearrangement network for visual
  inspection of textured surface defects toward edge intelligent
  manufacturing,'' {\em IEEE Transactions on Automation Science and
  Engineering}, 2022.

\bibitem{r2}
H.~Yang, Q.~Zhou, K.~Song, and Z.~Yin, ``An anomaly feature-editing-based
  adversarial network for texture defect visual inspection,'' {\em IEEE
  Transactions on Industrial Informatics}, vol.~17, no.~3, pp.~2220--2230,
  2020.

\bibitem{r3}
Q.~Wan, L.~Gao, X.~Li, and L.~Wen, ``Unsupervised image anomaly detection and
  segmentation based on pre-trained feature mapping,'' {\em IEEE Transactions
  on Industrial Informatics}, 2022.

\bibitem{r4}
X.~Tao, D.~Zhang, Z.~Wang, X.~Liu, H.~Zhang, and D.~Xu, ``Detection of power
  line insulator defects using aerial images analyzed with convolutional neural
  networks,'' {\em IEEE Transactions on Systems, Man, and Cybernetics:
  Systems}, vol.~50, no.~4, pp.~1486--1498, 2018.

\bibitem{r5}
H.~Dong, K.~Song, Y.~He, J.~Xu, Y.~Yan, and Q.~Meng, ``Pga-net: Pyramid feature
  fusion and global context attention network for automated surface defect
  detection,'' {\em IEEE Transactions on Industrial Informatics}, vol.~16,
  no.~12, pp.~7448--7458, 2020.

\bibitem{r6}
S.~Mei, H.~Yang, and Z.~Yin, ``An unsupervised-learning-based approach for
  automated defect inspection on textured surfaces,'' {\em IEEE Transactions on
  Instrumentation and Measurement}, vol.~67, no.~6, pp.~1266--1277, 2018.

\bibitem{r7}
J.~Yang, Y.~Shi, and Z.~Qi, ``Learning deep feature correspondence for
  unsupervised anomaly detection and segmentation,'' {\em Pattern Recognition},
  vol.~132, p.~108874, 2022.

\bibitem{r8}
A.~Dosovitskiy, L.~Beyer, A.~Kolesnikov, D.~Weissenborn, X.~Zhai,
  T.~Unterthiner, M.~Dehghani, M.~Minderer, G.~Heigold, S.~Gelly, {\em et~al.},
  ``An image is worth 16x16 words: Transformers for image recognition at
  scale,'' {\em arXiv preprint arXiv:2010.11929}, 2020.

\bibitem{r9}
D.~Gong, L.~Liu, V.~Le, B.~Saha, M.~R. Mansour, S.~Venkatesh, and A.~v.~d.
  Hengel, ``Memorizing normality to detect anomaly: Memory-augmented deep
  autoencoder for unsupervised anomaly detection,'' in {\em Proceedings of the
  IEEE/CVF International Conference on Computer Vision}, pp.~1705--1714, 2019.

\bibitem{r13}
P.~Bergmann, S.~L{\"o}we, M.~Fauser, D.~Sattlegger, and C.~Steger, ``Improving
  unsupervised defect segmentation by applying structural similarity to
  autoencoders,'' {\em arXiv preprint arXiv:1807.02011}, 2018.

\bibitem{r10}
S.~Akcay, A.~Atapour-Abarghouei, and T.~P. Breckon, ``Ganomaly: Semi-supervised
  anomaly detection via adversarial training,'' in {\em Asian conference on
  computer vision}, pp.~622--637, Springer, 2018.

\bibitem{r11}
T.~Schlegl, P.~Seeb{\"o}ck, S.~M. Waldstein, G.~Langs, and U.~Schmidt-Erfurth,
  ``f-anogan: Fast unsupervised anomaly detection with generative adversarial
  networks,'' {\em Medical image analysis}, vol.~54, pp.~30--44, 2019.

\bibitem{r12}
Y.~Yan, D.~Wang, G.~Zhou, and Q.~Chen, ``Unsupervised anomaly segmentation via
  multilevel image reconstruction and adaptive attention-level transition,''
  {\em IEEE Transactions on Instrumentation and Measurement}, vol.~70,
  pp.~1--12, 2021.

\bibitem{r16}
V.~Zavrtanik, M.~Kristan, and D.~Skočaj, ``Reconstruction by inpainting for
  visual anomaly detection,'' {\em Pattern Recognition}, vol.~112, p.~107706,
  2021.

\bibitem{r14}
Y.~Shi, J.~Yang, and Z.~Qi, ``Unsupervised anomaly segmentation via deep
  feature reconstruction,'' {\em Neurocomputing}, vol.~424, pp.~9--22, 2021.

\bibitem{r15}
X.~Tao, D.~Zhang, W.~Ma, Z.~Hou, Z.~Lu, and C.~Adak, ``Unsupervised anomaly
  detection for surface defects with dual-siamese network,'' {\em IEEE
  Transactions on Industrial Informatics}, vol.~18, no.~11, pp.~7707--7717,
  2022.

\bibitem{r17}
J.~Yang, Y.~Shi, and Z.~Qi, ``Learning deep feature correspondence for
  unsupervised anomaly detection and segmentation,'' {\em Pattern Recognition},
  vol.~132, p.~108874, 2022.

\bibitem{r18}
Q.~Wan, L.~Gao, X.~Li, and L.~Wen, ``Unsupervised image anomaly detection and
  segmentation based on pre-trained feature mapping,'' {\em IEEE Transactions
  on Industrial Informatics}, pp.~1--10, 2022.

\bibitem{r22}
K.~He, X.~Chen, S.~Xie, Y.~Li, P.~Dollár, and R.~Girshick, ``Masked
  autoencoders are scalable vision learners,'' in {\em 2022 IEEE/CVF Conference
  on Computer Vision and Pattern Recognition (CVPR)}, pp.~15979--15988, 2022.

\bibitem{r23}
J.~Pirnay and K.~Chai, ``Inpainting transformer for anomaly detection,'' in
  {\em International Conference on Image Analysis and Processing},
  pp.~394--406, Springer, 2022.

\bibitem{r19}
L.~Chen, Z.~You, N.~Zhang, J.~Xi, and X.~Le, ``Utrad: Anomaly detection and
  localization with u-transformer,'' {\em Neural Networks}, vol.~147,
  pp.~53--62, 2022.

\bibitem{r20}
J.~Jiang, J.~Zhu, M.~Bilal, Y.~Cui, N.~Kumar, R.~Dou, F.~Su, and X.~Xu,
  ``Masked swin transformer unet for industrial anomaly detection,'' {\em IEEE
  Transactions on Industrial Informatics}, 2022.

\bibitem{r21}
H.~{Yao}, X.~{Wang}, and W.~{Yu}, ``{Siamese Transition Masked Autoencoders as
  Uniform Unsupervised Visual Anomaly Detector},'' {\em arXiv e-prints},
  p.~arXiv:2211.00349, Nov. 2022.

\bibitem{r24}
P.~Bergmann, M.~Fauser, D.~Sattlegger, and C.~Steger, ``Mvtec ad—a
  comprehensive real-world dataset for unsupervised anomaly detection. in 2019
  ieee,'' in {\em CVF Conference on Computer Vision and Pattern Recognition
  (CVPR)}, pp.~9584--9592, 2020.

\bibitem{r25}
Y.~Lee and P.~Kang, ``Anovit: Unsupervised anomaly detection and localization
  with vision transformer-based encoder-decoder,'' {\em IEEE Access}, vol.~10,
  pp.~46717--46724, 2022.

\bibitem{r26}
J.~Gu, Z.~Wang, J.~Kuen, L.~Ma, A.~Shahroudy, B.~Shuai, T.~Liu, X.~Wang,
  G.~Wang, J.~Cai, and T.~Chen, ``Recent advances in convolutional neural
  networks,'' {\em Pattern Recognition}, vol.~77, pp.~354--377, 2018.

\end{thebibliography}
\end{document}